\documentclass[preprints,article,accept,moreauthors,pdftex]{Definitions/mdpi} 

\firstpage{1} 
\pubvolume{xx}
\issuenum{1}
\articlenumber{5}
\pubyear{2020}
\copyrightyear{2020}
\history{}
\Title{Towards Robust Word Embeddings for Noisy Texts}

\Author{Yerai Doval %Please carefully check the accuracy of names and affiliations.
 $^{1,}$*\orcidA{}, Jes\'us Vilares $^{2}$ and Carlos G\'omez-Rodr\'iguez $^{2}$\orcidC{}}

\AuthorNames{Yerai Doval, Jes\'us Vilares, and Carlos G\'omez-Rodr\'iguez}

\address{%
$^{1}$ \quad Grupo COLE, Escola Superior de Enxe\~{n}ar\'ia Inform\'atica, Universidade de Vigo, 36310 Vigo, Spain\\
$^{2}$ \quad Universidade da Coruña, CITIC. Grupo LyS, Departamento de Ciencias da Computaci\'on e Tecnolox\'ias da Informaci\'on,  15071 A Coru\~{n}a, %Please add post code. (or zip code in the US).
 Spain; jesus.vilares@udc.es (J.V.); carlos.gomez@udc.es (C.G.-R.)}

\corres{Correspondence: yerai.doval@uvigo.es}

\abstract{Research on word embeddings has mainly focused on improving their performance on standard corpora, disregarding the difficulties posed by noisy texts in the form of tweets and other types of non-standard writing from social media.
In this work, we propose a simple extension to the skipgram model in which we introduce the concept of bridge-words, which are artificial words added to the model to strengthen the similarity between standard words and their noisy variants.
Our new embeddings outperform baseline models on noisy texts on a wide range of evaluation tasks, both intrinsic and extrinsic, while retaining a good performance on standard texts.
To the best of our knowledge, this is the first explicit approach at dealing with these types of noisy texts at the word embedding level that goes beyond the support for out-of-vocabulary words.}


\begin{document}
%%%%%%%%%%%%%%%%%%%%%%%%%%%%%%%%%%%%%%%%%%

%%%%%%%%%%%%%%%%%%%%%%%%%%%%%%%%%%%%%%%%%%
%\setcounter{section}{-1} %% Remove this when starting to work on the template.
%\section{How to Use This Template}
%The template details the sections that can be used in a manuscript. Note that the order and names of article sections may differ from the requirements of the journal (e.g., the positioning of the Materials and Methods section). Please check the instructions for authors page of the journal to verify the correct order and names. For any questions, please contact the editorial office of the journal or support@mdpi.com. For LaTeX related questions please contact latex@mdpi.com.
%The order of the section titles is: Introduction, Materials and Methods, Results, Discussion, Conclusions for these journals: aerospace,algorithms,antibodies,antioxidants,atmosphere,axioms,biomedicines,carbon,crystals,designs,diagnostics,environments,fermentation,fluids,forests,fractalfract,informatics,information,inventions,jfmk,jrfm,lubricants,neonatalscreening,neuroglia,particles,pharmaceutics,polymers,processes,technologies,viruses,vision

\section{Introduction} 

Continuous word representations, also known as word embeddings, have been successfully used in a wide range of NLP tasks such as dependency parsing~\cite{Bansal2014}, information retrieval~\cite{vulic2015monolingual}, \mbox{POS tagging~\cite{Kutuzov2016},} or Sentiment Analysis (SA)~\cite{Xiong2018}. 
A popular scenario for NLP tasks these days is social media platforms such as Twitter~\cite{Lampos2017,Yang2018,Liang2018}, where texts are usually written without following the standard rules, containing varying levels of noise in the form of spelling mistakes (`socisl' for `social'), phonetic spelling of words (`dat' for `that'), abbreviations for common phrases (`tbh' for `to be honest'), emphasis (`yessss' as an emphatic `yes') or incorrect word segmentations (`noway' for `no way').
However, the most commonly-used word embedding approaches do not take these phenomena into account~\cite{Mikolov2013,Pennington2014,bojanowski2017enriching}, \mbox{and we} instead rely on their implicit capacity to cope with non-standard words provided a large enough amount of varied training text, such as in~\cite{Sumbler2018}.

Another possibility to tackle non-standard texts would be to apply some preprocessing step that removes the noise, such as spell checking or text normalization~\cite{Eisenstein2013,Chrupaa2014,Van2017b}.
Nonetheless, the trend nowadays is to use end-to-end approaches~\cite{Bordes2017,Klein2017,Schmitt2018} that exploit the raw data from the source without applying preprocessing steps, in an attempt to harness every bit of information for the specific task at hand while also avoiding introducing early errors in the NLP pipeline. 
On the other hand, it is also not entirely clear whether a normalization approach outperforms the direct use of word embeddings on noisy texts~\cite{Van2017a}.
Normalization, as a preprocessing step, will alter the original information encoded in the input text, although in a way that would benefit the next stages of the pipeline.
For instance, if we normalize `nooooo' to `no', the emphasis of the first word is lost. 
In this case, we should highlight the \emph{intentionality} when using one form over the other, which contrasts with accidentally introducing spelling mistakes in the writing which, nonetheless, may still convey some information such as the educational level of the writer. 
Granted, a system that only includes normalized words in its vocabulary will probably benefit from using the latter form instead.

In this work, we introduce an adaptation of the skipgram model from~\cite{bojanowski2017enriching} to train word embeddings that better integrate word variants (otherwise considered noisy words) at training time.
This can be regarded as an analogous incremental improvement over fastText to what this one was over word2vec.
Then, we perform an evaluation on a wide array of intrinsic and extrinsic tasks, comparing their performance to that of well-known embedding models such as word2vec and fastText on both standard and noisy English texts.
The results show a clear improvement over the baselines in semantic similarity and sentiment analysis tasks, with a general tendency to retain the performance of the best baseline on standard texts and outperform them on noisy texts.
Our ultimate goal is to improve the performance of traditional embedding models in the context of noisy texts.
This would alleviate the need for the usual preprocessing steps such as spell checking or microtext normalization, and act as \mbox{a good} starting point for modern end-to-end NLP approaches.

\section{Towards Noise-Resistant Word Embeddings}
\label{sec:towards}

Word embedding models such as word2vec, GloVe or fastText are able to cluster word variants together when given a big enough training corpus that includes standard and non-standard language~\cite{Sumbler2018}.
That is, given enough examples where `friend' (standard word), `freind' (spell-checking error), `frnd' (phonetic-compressed spelling) and even `dog' or `dawg' (street-talk) appear in similar contexts, these words will be translated to similar vector representations.
Taking advantage of this fact, many state-of-the-art microtext normalization systems use word embeddings in their pipelines~\cite{Bertaglia2017,Van2017b,Ansari2017,sridhar2015unsupervised}, both when generating normalization candidates for the input words and also when selecting them.

The problem with this approach is that the contexts where those example words appear are also likely to be affected by the same phenomena as the words themselves.
For example, `friend' might appear in phrases such as `that's my best friend' or `friend for life', while `frnd' in others such as `dats my bst frnd' or `frnd 4 lifee'.
This can make it difficult for the embedding algorithm to find the semantic similarity between `friend' and `frnd' when only relying on the assumption that the training corpus is big and diverse enough to effectively convey this variability.
However, not all of the embedding algorithms are equally affected by this, as those which take subword information into account may have an advantage: 
in our example, the similar morphology shared by the word variants may be exploited by algorithms such as fastText, which uses character n-grams to give them more similar vector representations. 

In this paper, we present a modification of the skipgram model proposed \mbox{by~\citet{bojanowski2017enriching}} (a modification of the original by~\citet{Mikolov2013}), which tries to improve the clustering of standard words and their noisy variants.
This is attained through the use of \emph{bridge-words}, normalized derivatives of the original words from the training corpus where one of their constituent characters is removed. 
By using these new words at training time in addition to the original ones, our objective is to increase the similarity between word variants, using those bridge-words as intermediate terms that match the words we want to cluster together. 
For example, `friend' and `freind' have in common the bridge-words `frind' and `frend'.
Even if the original words do not appear in the same context in the training corpus, using the bridge-words in place of the originals allows for indirect paths to be discovered: `friend'-`frind'-`freind' and `friend'-`frend'-`freind'.
In the case of `friend' and `frnd', and assuming that we use an embedding algorithm that exploits subword information, as we propose here, the higher morphological similarities of the latter with respect to the bridge-words `frend' and `frind' benefits their grouping together in the same cluster. 
As a side note, in the sense that these are intermediate (or normalized) representations that tie together otherwise isolated terms, they may resemble the index terms used in information retrieval. Since there is no index in our case, we will not refer to them as such.
\mbox{Notably, it should} also be possible to apply analogous modifications to the ones described here to other training models, such as the continuous bag of words~\cite{Mikolov2013}.

It is worth pointing out that we did not consider the latest state-of-the-art models such as BERT~\cite{devlin2018bert} and GPT-2~\cite{radford2019language} as it would not be feasible to apply analogous modifications to these large and complex models at this point, and GPT-3~\cite{brown2020language} is out of the question in this regard.
On the other hand, \mbox{although we} currently consider a monolingual English setup, our method should be suitable for any other language with a similar concept of \textit{character}, in contrast to those based on logograms such \mbox{as Chinese.}

\subsection{Modified Skipgram Model}
\label{subsection:modified-skipgram-model}

The skipgram model found in tools like word2vec and fastText establishes that, for each word in \mbox{a text}, it should be possible to predict those in their corresponding contexts~\cite{Mikolov2013}.
\mbox{As a consequence,} \mbox{the words} that appear in similar contexts end up represented by similar vectors, so that the transformation learned by the model can effectively map one group of words onto the other.

\begin{figure}[H] 
    \centering
    \includegraphics[width=.5\columnwidth]{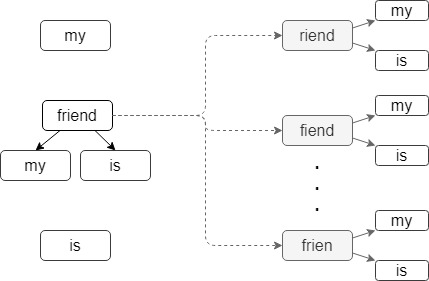} 
  \caption{Visualization of the adapted skipgram model where we add bridge-words into its training.}
  \label{fig:skipgram} 
\end{figure}

Based on the skipgram model from fastText, our proposal aims at increasing the similarity between standard words and their noisy counterparts.
To accomplish this, we introduce a new set of words at training time that we denominate \emph{bridge-words}.
For each word in the training corpus, we first put the words into lowercase, strip the accents and remove character successive repetitions, and then obtain 
one bridge-word for each remaining character in the word,
%$n$ bridge-words 
by removing one different character each time.
For character repetitions, we cover both standard and non-standard ones (e.g., `success' vs `daaammn'), obtaining a common denominator for when users make the mistake of removing standard repetitions (e.g., from `success' to `succes'), or add repetitions to provide emphasis (e.g., from `damn' to `daaammn').
The resulting words are very similar and can still be read mostly in the same way.
An analogous reasoning is used in the case of lowercasing and stripping the accents.
Note that this procedure is exclusively applied to obtain all the bridge-words, and the unprocessed corpus will be used during training.
Formally, let $\mathcal{V}$ be the word vocabulary extracted from the training corpus so that $\mathcal{V} = \{w_1, w_2, ..., w_n\}$ with $n$ the size of the vocabulary.
The set of bridge-words is then defined as $\mathcal{B} = \{b_{1,1}, b_{1,2}, ..., b_{1,|w_{1}|}, ..., b_{n,1}, b_{n,2}, ..., b_{n,|w_{n}|}\}$, where $|w_i|$ is the length of word $w_i$, and $b_{i,j}$ is the bridge-word obtained by first normalizing as described earlier and then removing the character at position $j$ from the word $w_i \in \mathcal{V}$ (it is possible that $\mathcal{V} \cap \mathcal{B} \neq \emptyset$).
These new words are used in addition to the original words when predicting their context in the skipgram model training, as depicted in Figure~\ref{fig:skipgram}.
For example, in the phrase `that's my best Fri\``endd ever', the objective is not only to predict `that's', `my', `best' and `ever' using the word `Fri\''endd', but also using the derived bridge-words `riend', `fiend', `frend', `frind', \mbox{`fried' and `frien'.}
This idea of removing one character at a time is similar to the one used in the tool SymSpell (\url{https://github.com/wolfgarbe/SymSpell}) to speed up spell-checking, where it replaces the exhaustive approach of considering all possible edit operations (i.e., addition, removal, substitution and transposition).
In our case, bridge-words are not interesting \textit{per se} but as \textit{intermediaries} between other words.
\mbox{We do not} require that they coincide with real words with which they would establish a direct connection; in fact, we assume that these connections will be indirect most of the time.
For instance, we do not consider the substitution operations that would construct `tome' and `tame' from `time', which would explicitly connect the three, but only `tme', which can be obtained from the three of them by removing one character, linking them together indirectly.

It is important to observe that these bridge-words also constitute artificial noise introduced in our training process that could play a harmful role.
As an example, the word `fiend' appears as a bridge-word for `friend', while also being a standard word from the English dictionary without much semantic relation to the concept of friendship.
Because of this, bridge-words should not have the same impact as the original words when tuning the parameters of the model.
We propose two mechanisms for lowering the weight of bridge-words in the training process: (1) introducing them randomly, with a fixed probability $p_b$, instead of for all the original words, and (2) reducing the impact in the objective function by adding a weighting factor.
Formally, let $w_x$ be an input word of length $|w_x|$, $b_{j}$ the bridge-word for $w_x$ when the character at position $j$ is removed, $w_y$ a target word in the context of $w_x$, $H$ a random variable with $P(H=1) = p_b$ and $P(H=0) = 1 - p_b$, $h \sim H$, $\lambda$ the weight factor and $E_{ft}(w_x, w_y)$ the objective function of the skipgram model from fastText, then our new objective function, $E_{robust}$ is defined as:
\[ 
E_{robust} = E_{ft}(\mathbf{w_x}, \mathbf{w_y}) + h \cdot \lambda \cdot \sum_{j=1}^{|w_x|} E_{ft}(\mathbf{b_{j}}, \mathbf{w_y})
\]
where $\mathbf{w_x}$, $\mathbf{w_y}$, and $\mathbf{b_j}$ are the vector representations of the corresponding input, target and bridge-words.

In any case, the proposed technique does not rule out the requirement of a training corpus where standard and noisy variants of words are used.
Rather, it enhances the capacity of already existing models (in this case, the skipgram model from fastText) to \emph{bridge} or further interconnect these word variants. The corresponding source code is available at \url{https://github.com/yeraidm/bridge2vec}. 

\section{Evaluation}

We use multiple intrinsic and extrinsic evaluation tasks to study the performance of our approach together with word2vec and fastText. 
The models are trained using the same unprocessed corpus of web text and tweets.
Starting with the usual word similarity task, we also include outlier detection~\cite{camacho2016find}, most of the extrinsic tasks from the SentEval benchmark~\cite{conneau2018senteval}, and then we add Twitter SA from various editions of the SemEval workshop.
Ideally, we should see that our embeddings are able to retain the performance of `vanilla' fastText embeddings~\cite{bojanowski2017enriching} for standard and less-corrupted text, \mbox{while outperforming} them on noisier texts, and that word2vec~\cite{Mikolov2013} is at a disadvantage in this case.

It is worth noting that including models like BERT in our benchmarks would be unfair given their significantly higher complexity and the amount of resources employed in their training with respect to the current ones which, sadly, are out of our reach. 
Our initial aim was to improve upon existing easy-to-train models (much more affordable to researchers), leaving the latest language models out of the question for this work. After all, we are not presenting a totally new embedding model but \mbox{a technique} to enhance existing ones. 

We believe that fastText and word2vec remain an accessible way to obtain competent embeddings in many scenarios, including for low-resource languages where bigger models would need more training data.
In addition, this is not to mention the significantly higher amount of computational resources required by these models in general.

In any case, it would be interesting (although out of scope for this work) to confirm whether BERT and similar models work well on noisy texts, which would require testing them against comparable models that take this use-case more explicitly, as we do in our paper for smaller embedding models. 

\subsection{Word Embedding Training}

In the present work, we use a combination of web corpora, specifically the UMBC corpus~\cite{UMBC}, and tweets collected through the Twitter Streaming API from dates between October 2015 and July 2018.
It is worth noting that we did not perform any preprocessing or normalization step over the resulting corpus, 
and the final dataset is formed by 64.653 M lines and 3.3 B tokens, of which 24.558 M are unique.

We employed a modified version of the skipgram model from fastText which incorporates the changes described in Section~\ref{subsection:modified-skipgram-model} together with a vanilla version (\url{https://github.com/facebookresearch/fastText}) and a word2vec baseline (\url{https://github.com/dav/word2vec}), using the default hyperparameters for all models.
\mbox{In the case} of the proposed model, we train four instances in order to take a first look at the influence of the hyperparameters introduced: the probability of introducing a bridge-word ($p_b$) and the weight for bridge-words in the objective ($\lambda$).
The combinations are $(p_b=1, \lambda=1)$, $(p_b=0.5, \lambda=1)$, $(p_b=1, \lambda=0.1)$, and $(p_b=0.5, \lambda=0.1)$.
In this work, we do not perform hyperparameter optimization given resource and time constraints, and those values were selected according to the initial hypothesis that a decreased impact of bridge-words in the training process should be beneficial to the model.
The training of our models is four times slower than vanilla fastText and word2vec when $p_b=0.5$ and 6.5 times slower when $p_b=1$ on average. 

\subsection{Intrinsic Tasks: Word Similarity and Outlier Detection}

The first intrinsic evaluation task is the well-known semantic word similarity task.
It consists of scoring the similarity between pairs of words, and comparing it to a gold standard given by human annotators.
In a word embedding space, the similarity between two words can be measured through \mbox{a distance} or similarity metric between the corresponding vectors in the space, such as cosine similarity.
The evaluation is performed using the Spearman correlation between the list of similarity scores obtained and the gold standard.
In this work, we use the wordsim353~\cite{Levetal:2002}, SCWS~\cite{huang2012improving}, SimLex999~\cite{hill2015simlex}, and SemEval17 (monolingual)~\cite{semeval2017similarity} evaluation datasets.

The second task is outlier detection, which consists of identifying the word that does not belong in a group of words according to their pair-wise semantic similarities.
As an example, snake would be \mbox{an outlier} in the set german shepherd, golden retriever, and french pitbull, in spite of also being an animal, since it is not a dog.
In this case, we use the 8-8-8~\cite{camacho2016find} and wiki-sem-500~\cite{blair2016automated} datasets, \mbox{and measure} the proportion of times in which the outlier was successfully detected (i.e., the accuracy).

\subsection{Extrinsic Tasks: The Senteval Benchmark and Twitter SA}
\label{ssec:extrinsic}

Since it is not evident that performance on intrinsic tasks translates proportionally to extrinsic tasks~\cite{faruqui2016problems,chiu2016intrinsic}, where word embeddings are used as part of bigger systems, we resort to the SentEval benchmark~\cite{conneau2018senteval} in order to evaluate our embeddings in a more realistic setup.
The tasks included in this benchmark evaluate sentence embeddings, which can be obtained from word embeddings using an aggregating function, which can go from the simple bag of words to the more complex neural-based models InferSent~\cite{conneau2017supervised} or GenSen~\cite{subramanian2018learning}.
Additionally, some tasks require a classifier to be trained on the sentence embeddings in order to obtain an output of the desired type.
In both cases, \mbox{we maintain} a simple approach where we focus on the raw performance of the word embeddings rather than the models used on top of them.
This means using the bag of words model to obtain sentence representations, which simply averages the corresponding word embeddings from each sentence, and then linear regression for the classification tasks.

SentEval includes 17 extrinsic tasks, of which we use 16 and 10 probing tasks.
The first group includes semantic textual similarity (STS 2012-2016, STS Benchmark and SICK-Relatedness), natural language inference (SICK-Entailment and SNLI), sentiment analysis (SST, both binary and fine-grained), opinion--polarity (MPQA), movie and product review (MR and CR), subjectivity status (SUBJ), question-type classification (TREC), and paraphrase detection (MRPC).
The second group is formed by tasks that evaluate other linguistic properties which could be found encoded in sentence embeddings, such as sentence length, depth of the syntactic tree or the number of the subject of the main clause.
For a more detailed description of these tasks together with references to the original sources, see~\cite{conneau2018senteval}.
In general, for the similarity tasks, the performance is measured using Spearman correlation, while, in the rest of the cases, which correspond to classification tasks, the accuracy of the classification is obtained.
Unfortunately, we leave image-caption retrieval task (COCO) out of our test bench as it is not possible to access the source texts.
This would be needed for the processing that we perform as described in the next section.

Finally, we also evaluate on the SA datasets released in the SemEval workshops \mbox{by \citet{nakov2013semeval}} (task 2, subtask B), \citet{rosenthal2014semeval} (task 9, subtask B, using the training data from the previous edition), and \citet{nakov2016semeval} (task 4, subtasks B, D, C, and E).
\mbox{These already} include noisy texts in the form of tweets, thus they are not processed in the same way as the following datasets are processed, as explained below.
However, since we still use the SentEval code, we did filter the neutral/objective tweets in ternary SA datasets.
We also performed downsampling on the 2016 training and development datasets, both binary and fine-grained, in order to compensate for the substantial unbalance across instance classes. 
This is important as the test datasets are also skewed in the same manner, and it lead the classifiers to adjust to this bias to obtain unrealistic results.
In the case of the binary task, we equated the positive instances with the number of negative ones, while, in the case of the fine-grained task, we used a fixed maximum number of 500 instances per class, \mbox{given the} huge gap between the least frequent class (accounting for 71 instances) and the most frequent one (including 2876 instances).
Note that other datasets used in this work are also unbalanced, although to \mbox{a significantly} lesser extent and with no such measurable impact on the results.

\subsection{Dataset De-Normalization}

Since we could not find noisy text datasets for such a wide variety of evaluation tasks as the ones from the SentEval benchmark, 
we decided to \emph{de-normalize} (i.e., introduce artificial noise into) these standard datasets, while also keeping the originals of the benchmark, in order to cover the case of noisy texts in the extension needed by this work.
The procedure consists of randomly replacing every word in the texts with a noisy variant with some fixed probability.
The noisy variants are obtained from two publicly available normalization dictionaries, \texttt{utdallas} and \texttt{unimelb}, released in the first edition (2015) of the W-NUT workshop~\cite{baldwin2015shared}, formed by (non-standard, standard) word pairs.

For the word similarity and outlier detection datasets, this probability $p_d$ was fixed to 1; \mbox{i.e., we modify} all the words in the test set which appear in our normalization dictionaries (which cover 78.61\% of them).
In the case of the SentEval datasets, we created three versions for each one of them: a heavily corrupted version ($p_d=1$), a more balanced version ($p_d=0.6$), and a less noisy one ($p_d=0.3$).
As an example, from the original sentence `A man is playing a flute,' we obtain `aa woma isz playiin thw flute', `aa mann is playng da flute', and `aa wman is playing the flute', in each respective case.
The Twitter SA datasets, on the other hand, were not de-normalized.

Furthermore, we perform ten de-normalization runs over the intrinsic tasks datasets and three over the extrinsic ones, obtaining multiple noisy versions of each dataset.
By averaging the results over the different de-normalizations, we try to neutralize extreme measurements that can be caused by different noisy variants of words.

\subsection{Results}

Our currently best model is obtained with the hyperparameter combination \mbox{$(p_b=0.5, \lambda=0.1)$}, which in some way validates our hypothesis that bridge-words should be introduced in a restrained fashion.
In general terms, this model has a similar performance to fastText in the standard case, \mbox{while outperforming} both word2vec and fastText in noisy setups, with wider margins towards \mbox{noisier texts.}

\subsubsection{Intrinsic Evaluation} 

Table~\ref{table:word_similarity} shows the results on the intrinsic word similarity task.
On standard words, fastText and our model obtain similar performance, both surpassing that of word2vec.
On non-standard words, however, our model is able to consistently outperform fastText in every dataset, and word2vec falls further behind possibly due to its lack of support for out-of-vocabulary words in this scenario, \mbox{as 48.77\%} of the unique noisy test words are not included in the vocabulary of the word2vec model.
Differences for non-standard words between our model and both word2vec and fastText are statistically significant under a significance level of 0.01.
\begin{table}[H]
\centering
%\small
\caption{Spearman correlation results of word similarity on SCWS, wordsim353 (WS353), SimLex999 (SL999), and SemEval17 (Sem17) datasets.}
\begin{tabular}{ccccc}
                              \toprule \multicolumn{5}{c}{\textbf{Standard}} 
\\
                              
                              \multicolumn{1}{c}{} & \multicolumn{1}{c}{\textbf{SCWS}} & \multicolumn{1}{c}{\textbf{WS353}} & \multicolumn{1}{c}{\textbf{SL999}} & \multicolumn{1}{c}{\textbf{Sem17}} \\ \cmidrule{1-5} 
\multicolumn{1}{c}{word2vec}      & 64.7                    & 69.1                          & 32.2                         & 68.2                         \\
\multicolumn{1}{c}{fastText} & \textbf{65.4} 
                   & 72.7                          & 33.5                         & 70.3                         \\
\multicolumn{1}{c}{ours}     & 65.1                    & \textbf{73.1}                          & \textbf{33.8}                         & \textbf{70.4}                         \\
                              \midrule
                               \multicolumn{5}{c}{\textbf{Noisy}}\\
                              
                              \multicolumn{1}{c}{} & \multicolumn{1}{c}{\textbf{SCWS}} & \multicolumn{1}{c}{\textbf{WS353}} & \multicolumn{1}{c}{\textbf{SL999}} & \multicolumn{1}{c}{\textbf{Sem17}} \\ \cmidrule{1-5} 
\multicolumn{1}{c}{word2vec}      & 13.0                    & 13.7                          & $-$10.9                        & 11.2                         \\
\multicolumn{1}{c}{fastText} & 35.2                    & 38.1                          & 7.3                          & 37.2                         \\
\multicolumn{1}{c}{ours}     & \textbf{42.1}                    & \textbf{44.2}                          & \textbf{16.4}                         & \textbf{43.1} \\
\bottomrule
\end{tabular}
\label{table:word_similarity}
\end{table}

In the case of outlier detection, shown in Table~\ref{table:outlier_detection}, we obtained mixed results and the differences between our model and the baselines are not statistically significant.
On the 8-8-8 dataset, \mbox{our model} outperforms the baselines both in the standard and noisy scenarios, although with visibly lower margins than in the case of semantic similarity.
However, on the wiki-sem-500 dataset, \mbox{word2vec outperforms} its competitors on standard words and does not lose much performance on the noisy setup.
The latter may be explained by the low amount of successfully denormalized words, with just 7.5\% of the total (compared to 52.2\% on the 8-8-8 dataset), which also hints at the tie between fastText and our model.

\begin{table}[H]
\centering
%\small
\caption{Accuracy results of outlier detection on 8-8-8 and wiki-sem-500 (wiki) datasets.}
\begin{tabular}{ccccc}
                                \toprule
                              & \multicolumn{2}{c}{\textbf{Standard}}                             & \multicolumn{2}{c}{\textbf{Noisy}}                        \\ 
\multicolumn{1}{c}{}         & \multicolumn{1}{c}{\textbf{8-8-8}} & \multicolumn{1}{c}{\textbf{wiki}} & \multicolumn{1}{c}{\textbf{8-8-8}} & \multicolumn{1}{c}{\textbf{wiki}} \\ \midrule
\multicolumn{1}{c}{word2vec}      & 59.4                         & \textbf{53.8}                     & 22.8                         & 39.3                    \\
\multicolumn{1}{c}{fastText} & 65.6                         & 49.0                     & 31.7                         & \textbf{41.1}                    \\
\multicolumn{1}{c}{ours} & \textbf{67.2}                         & 47.8                     & \textbf{33.3}                         & \textbf{41.1}    \\
\bottomrule
\end{tabular}
\label{table:outlier_detection}
\end{table}

\subsubsection{Extrinsic Evaluation} 

Given the considerable amount of tasks and datasets included in the SentEval benchmark, \mbox{we decided} to group similar tasks and datasets and show the aggregated results from each group instead of following an exhaustive approach. 
In this case, and given the variability in dataset sizes, \mbox{we use} a weighted average as the aggregation function.

\begin{figure}[H] 
    \centering
    \includegraphics[width=.65\columnwidth]{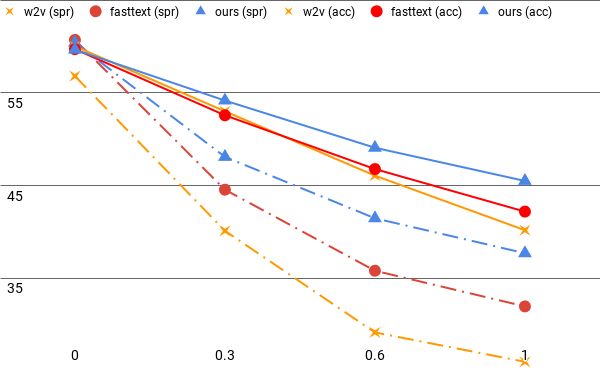} 
  \caption{Performance of each considered model when going from standard texts to noisier ones on the extrinsic tasks.
  In lines and dots is the aggregated performance on semantic similarity and relatedness tasks (Spearman correlation).
  In continuous lines is the aggregated performance on the rest of the \mbox{tasks (accuracy).}}
  \label{fig:trend} 
\end{figure}

First of all, we show in Figure~\ref{fig:trend} the dynamic behavior of each model when going from standard texts to noisier ones.
In this case, we divided the tasks into two groups based on the performance metric: Spearman correlation or accuracy. 
The first one encompasses the semantic similarity and relatedness tasks (all STS* and SICK-Relatedness) and the second one the rest of them.
Except in the case of word2vec on the first group (yellow lines and crosses), all the models start from a very similar position in the standard scenario.
Then, the performance begins its downward trend, where our model starts to stand out above the baselines.
As we go towards noisier texts, our model manages to stay above the rest of the lines, increasing the distance margin up until the last stretch.

Next, Table~\ref{table:senteval} shows in greater detail the performance of each model in a less aggregated view.
\mbox{In this case,} datasets have been grouped by task as described in Section~\ref{ssec:extrinsic}.
As we can see, our model is on par with the baselines on standard texts, with a few interesting exceptions: (1) it is able to obtain some advantage on sentiment analysis, which fastText also obtains over word2vec; (2) on question-type classification, word2vec obtains the best performance, and still clearly outperforms fastText on the lowest noise level, although not our model; and (3) on the probing tasks, word2vec takes the lead again, this time by a smaller margin.
Regarding noisy texts, our model is clearly superior on semantic similarity and relatedness, as we had already seen before, and it also outperforms the baselines on the rest of the tasks, with wider margins on noisier texts, but with the sole exception of paraphrase detection.
In this surprising case, word2vec outperforms both fastText and our model obtaining better accuracy on texts with the highest level of noise compared to the previous step. 
It appears that, with the proper training (and hence, vocabulary), word2vec remains a strong baseline on extrinsic tasks, even in the case of noisy texts, where the level of noise has to be increased notably in order for fastText to obtain a clear advantage.
This can also be observed following the continuous lines in Figure~\ref{fig:trend}.
\mbox{On the other hand,} the weakness seen on word semantic similarity (Table~\ref{table:word_similarity}) relating to out-of-vocabulary words does not seem to translate to extrinsic tasks, where having more context and hence a higher chance of finding in-vocabulary words mitigates the problem, as we can see in the semantic similarity and relatedness (Table~\ref{table:senteval}) results.
In any case, differences on noisy texts between our model and the baselines are statistically significant under a significance level of 0.05, with \emph{p}-values below or barely above 0.01.

\begin{table}[H]
\centering
\caption{Results of the extrinsic evaluation on the SentEval benchmark. The noise levels are \textit{low} ($p_d=0.3$), \textit{mid} ($p_d=0.6$), and \textit{high} ($p_d=1$).}
%\small
\bgroup
\def\arraystretch{1.15}
\scalebox{1}{
\begin{tabular}{ccccccccc}
   \toprule
    \multicolumn{1}{c}{}         & \multirow{2}{*}{\textbf{Standard}} & \multicolumn{3}{c}{\textbf{Noisy}}  & \multirow{2}{*}{\textbf{Standard}} & \multicolumn{3}{c}{\textbf{Noisy}}                                                     %& \multicolumn{1}{c}{\multirow{2}{*}{}} & \multicolumn{3}{c}{}                                               
    \\ 
    \multicolumn{1}{c}{}         &                           & \multicolumn{1}{c}{\textit{\textbf{Low}}} & \multicolumn{1}{c}{\textit{\textbf{Mid}}} & \multicolumn{1}{c}{\textit{\textbf{High}}} & & \multicolumn{1}{c}{\textit{\textbf{Low}}} & \multicolumn{1}{c}{\textit{\textbf{Mid}}} & \multicolumn{1}{c}{\textit{\textbf{High}}}%& \multicolumn{1}{c}{}                  & \multicolumn{1}{c}{} & \multicolumn{1}{c}{} & \multicolumn{1}{c}{} 
    \\ \midrule%\cline{1-5}

                              %\hline
                              & \multicolumn{4}{c}{\textbf{Semantic sim. \& rel.}} & \multicolumn{4}{c}{\textbf{Binary classification}}                                                    %& \multicolumn{4}{c}{}                                                                                       
                              \\ \midrule
\multicolumn{1}{c}{word2vec}      & \multicolumn{1}{c}{56.8}  & 40.1                    & 29.2                       & 26.0       & \multicolumn{1}{c}{81.5} & 78.8                   & 76.1                      & 71.4              %& \multicolumn{1}{r}{}                  & \multicolumn{1}{r}{} & \multicolumn{1}{r}{} & \multicolumn{1}{r}{} 
\\
\multicolumn{1}{c}{fastText} & \multicolumn{1}{c}{\textbf{60.7}}  & 44.6                    & 35.8                       & 32.0  & \multicolumn{1}{c}{\textbf{81.8}} & 79.0                   & 75.8                      & 72.1                   %& \multicolumn{1}{r}{}                  & \multicolumn{1}{r}{} & \multicolumn{1}{r}{} & \multicolumn{1}{r}{} 
\\
\multicolumn{1}{c}{ours}     & \multicolumn{1}{c}{60.4}  & \textbf{48.1}                    & \textbf{41.5}                       & \textbf{37.7}   & \multicolumn{1}{c}{81.6} & \textbf{79.4}                   & \textbf{77.7}                      & \textbf{74.1}                  %& \multicolumn{1}{r}{}                  & \multicolumn{1}{r}{} & \multicolumn{1}{r}{} & \multicolumn{1}{r}{} 
\\
                              %& \multicolumn{1}{l}{}      & \multicolumn{1}{l}{}    & \multicolumn{1}{l}{}       & \multicolumn{1}{l}{}     &                                       &                      &                      &                      \\
                              %\hline
                                                                                                  %&                                       &                      &                      &                      
                              %\\ \hline
%\multicolumn{1}{l|}{}         & \multirow{2}{*}{\textit{standard}} & \multicolumn{3}{c}{\textit{noisy}}                                                       %&                                       &                      &                      &                      
%\\
%\multicolumn{1}{l|}{}         &                           & \multicolumn{1}{c}{low} & \multicolumn{1}{c}{mid} & \multicolumn{1}{c}{high} %&                                       &                      &                      &                      
%\\ \cline{1-5}
                    %&                                       &                      &                      &                      
%\\
                              %& \multicolumn{1}{l}{}      & \multicolumn{1}{l}{}    & \multicolumn{1}{l}{}       & \multicolumn{1}{l}{}     &                                       &                      &                      &                      \\
                              \hline & \multicolumn{4}{c}{\textbf{Sentiment analysis}} & \multicolumn{4}{c}{\textbf{Entailment}}                                                                      %& \multicolumn{1}{c}{}                  &                      &                      &                      
                              \\ \midrule
%\multicolumn{1}{l|}{}         & \multirow{2}{*}{\textit{standard}} & \multicolumn{3}{c}{\textit{noisy}} & \multirow{2}{*}{\textit{standard}} & \multicolumn{3}{c}{\textit{noisy}}                                                     %& \multicolumn{1}{c}{\multirow{2}{*}{}} & \multicolumn{3}{c}{}                                               
%\\
%\multicolumn{1}{l|}{}         &                           & \multicolumn{1}{c}{low} & \multicolumn{1}{c}{mid} & \multicolumn{1}{c}{high} & & \multicolumn{1}{c}{low} & \multicolumn{1}{c}{mid} & \multicolumn{1}{c}{high} %& \multicolumn{1}{c}{}                  & \multicolumn{1}{c}{} & \multicolumn{1}{c}{} & \multicolumn{1}{c}{} 
%\\ \hline%\cline{1-5}
\multicolumn{1}{c}{word2vec}      & \multicolumn{1}{c}{57.9} & 55.4                   & 50.5                      & 42.8 & \multicolumn{1}{c}{\textbf{66.3}} & 54.9                   & 48.8                      & 35.8                   %& \multicolumn{1}{r}{}                  & \multicolumn{1}{r}{} & \multicolumn{1}{r}{} & \multicolumn{1}{r}{} 
\\
\multicolumn{1}{c}{fastText} & \multicolumn{1}{c}{58.8} & 55.8                   & 52.6                      & 47.0    & \multicolumn{1}{c}{66.2} & 54.0                   & 50.0                      & 40.2                %& \multicolumn{1}{r}{}                  & \multicolumn{1}{r}{} & \multicolumn{1}{r}{} & \multicolumn{1}{r}{} 
\\
\multicolumn{1}{c}{ours}     & \multicolumn{1}{c}{\textbf{59.3}} & \textbf{56.9}                   & \textbf{54.6}                      & \textbf{51.1}   & \multicolumn{1}{c}{\textbf{66.3}} & \textbf{55.1}                   & \textbf{51.4}                      & \textbf{48.1}                 %& \multicolumn{1}{r}{}                  & \multicolumn{1}{r}{} & \multicolumn{1}{r}{} & \multicolumn{1}{r}{} 
\\
                              %& \multicolumn{1}{l}{}      & \multicolumn{1}{l}{}    & \multicolumn{1}{l}{}       & \multicolumn{1}{l}{}     &                                       &                      &                      &                      \\
                              %\hline                                                                              %&                                       &                      &                      &                      
                              
%\multicolumn{1}{l|}{}         & \multirow{2}{*}{\textit{standard}} & \multicolumn{3}{c}{\textit{noisy}}                                                      % &                                       &                      &                      &                      
%\\
%\multicolumn{1}{l|}{}         &                           & %\multicolumn{1}{c}{low} & \multicolumn{1}{c}{mid} & \multicolumn{1}{c}{high} %&                                       &                      &                      &                      
%\\ \cline{1-5}

                              %& \multicolumn{1}{l}{}      & \multicolumn{1}{l}{}    & \multicolumn{1}{l}{}       & \multicolumn{1}{l}{}     &                                       &                      &                      &                      \\
                              \hline & \multicolumn{4}{c}{\textbf{Question-type classification}}  & \multicolumn{4}{c}{\textbf{Paraphrase detection}}                                                         %& \multicolumn{1}{c}{}                  &                      &                      &                      
                              \\ \midrule
%\multicolumn{1}{l|}{}         & \multirow{2}{*}{\textit{standard}} & \multicolumn{3}{c}{\textit{noisy}}  & \multirow{2}{*}{\textit{standard}} & \multicolumn{3}{c}{\textit{noisy}}                                                    % & \multicolumn{1}{c}{\multirow{2}{*}{}} & \multicolumn{3}{c}{}                                               
%\\
%\multicolumn{1}{l|}{}         &                           & \multicolumn{1}{c}{low} & \multicolumn{1}{c}{mid} & \multicolumn{1}{c}{high} & & \multicolumn{1}{c}{low} & \multicolumn{1}{c}{mid} & \multicolumn{1}{c}{high} %& \multicolumn{1}{c}{}                  & \multicolumn{1}{c}{} & \multicolumn{1}{c}{} & \multicolumn{1}{c}{} 
%\\ \hline %\cline{1-5}
\multicolumn{1}{c}{word2vec}      & \multicolumn{1}{c}{\textbf{79.4}} & 65.6                   & 53.3                      & 35.0   & \multicolumn{1}{c}{72.6} & \textbf{67.0}                   & \textbf{61.2}                      & \textbf{65.5}                 %& \multicolumn{1}{r}{}                  & \multicolumn{1}{r}{} & \multicolumn{1}{r}{} & \multicolumn{1}{r}{} 
\\
\multicolumn{1}{c}{fastText} & \multicolumn{1}{c}{74.8} & 62.5                   & 52.1                      & 41.8    & \multicolumn{1}{c}{72.3} & 62.7                   & 57.1                      & 56.8                %& \multicolumn{1}{r}{}                  & \multicolumn{1}{r}{} & \multicolumn{1}{r}{} & \multicolumn{1}{r}{} 
\\
\multicolumn{1}{c}{ours}     & \multicolumn{1}{c}{73.4} & \textbf{67.5}                   & \textbf{59.4}                      & \textbf{49.5}  & \multicolumn{1}{c}{\textbf{72.9}} & 66.9                   & 60.2                      & 56.3                  %& \multicolumn{1}{r}{}                  & \multicolumn{1}{r}{} & \multicolumn{1}{r}{} & \multicolumn{1}{r}{} 
\\
                              %& \multicolumn{1}{l}{}      & \multicolumn{1}{l}{}    & \multicolumn{1}{l}{}       & \multicolumn{1}{l}{}     &                                       &                      &                      &                      \\
                              %\hline                                                                     %&                                       &                      &                      &                  
                    %&                                       &                      &                      &                      
                                       %& \multicolumn{1}{c}{}                  &                      &                      &                      
                         
                              %& \multicolumn{1}{l}{}      & \multicolumn{1}{l}{}    & \multicolumn{1}{l}{}       & \multicolumn{1}{l}{}     &                                       &                      &                      &                      \\
                              \hline & \multicolumn{4}{c}{\textbf{Probing tasks}}                  &                                       &                      &                      &       \\ \midrule              
                             % \\ %\cline{1-5}
%\multicolumn{1}{l|}{}         & \multirow{2}{*}{\textit{standard}} & \multicolumn{3}{c}{\textit{noisy}}                                                       %&                                       &                      &                      &                      
%\\
%\multicolumn{1}{l|}{}         &                           & \multicolumn{1}{c}{low} & \multicolumn{1}{c}{mid} & \multicolumn{1}{c}{high} %&                                       &                      &                      &                      
%\\ \cline{1-5}
\multicolumn{1}{c}{word2vec}      & \multicolumn{1}{c}{\textbf{58.2}} & 51.5                   & 45.3                      & \multicolumn{1}{r}{39.3}                    &                                       &                      &                      &                      
\\
\multicolumn{1}{c}{fastText} & \multicolumn{1}{c}{57.8} & 51.2                   & 45.9                      & \multicolumn{1}{c}{41.0}                    &                                       &                      &                      &                      
\\
\multicolumn{1}{c}{ours}     & \multicolumn{1}{c}{57.7} & \textbf{52.8}                   & \textbf{48.4}                      & \multicolumn{1}{c}{\textbf{43.6}}                  &                                       &                      &                      &                     \\\bottomrule
\end{tabular}}
\egroup
\label{table:senteval}
\end{table}

Finally, in Table~\ref{table:semeval}, we show the results obtained on the SemEval Twitter SA datasets.
In this case, word2vec continues to display a strong performance, fastText loses the advantage it had on the SentEval benchmark for the same SA task, and our approach is able to revert this performance loss to outperform, once again, both of the baselines.
Performance differences are statistically significant under a significance level of 0.05, again with $p$-values below or barely above 0.01.
At this point, we can observe how fastText is inferior to word2vec on a real-world social media setting, when we may have expected the opposite at first.
However, for this same reason, it is remarkable to see our approach taking the lead despite being a modification of fastText, which also demonstrates the benefit of including the bridge-words at training time.
Having said that, it would be relevant to investigate if higher performance figures can be obtained by modifying the skipgram model from word2vec.

\begin{table}[H]
%\small
\centering
\caption{Accuracy results of the extrinsic evaluation on SemEval (SE) Twitter SA datasets.}
\begin{tabular}{ccccc}
\toprule
         & \multicolumn{1}{c}{\textbf{SE13 B}} & \multicolumn{1}{c}{\textbf{SE14 B}} & \multicolumn{1}{c}{\textbf{SE16 BD}} & \multicolumn{1}{c}{\textbf{SE16 CE}} \\ \midrule
word2vec      & 84.3                           & 88.3                           & 77.4                            & 35.1                            \\
fastText & 83.3                           & 88.1                           & 76.5                            & 33.7                            \\
ours     & \textbf{84.8}                           & \textbf{88.6}                            & \textbf{78.4}                            & \textbf{35.5}     \\
\bottomrule
\end{tabular}
\label{table:semeval}
\end{table}

\section{Related Work}
\label{sec:related}

Word embeddings have been at the forefront of NLP research for the past decade, although the first application of vector representation of words dates back to~\cite{Rumelhart1986}.
More recently, the first models to attain wide use were word2vec~\cite{Mikolov2013} and GloVe~\cite{Pennington2014}, which take words as basic and indivisible units, implying that the word vocabulary is fixed at training time and any unknown word would be given the same vector representation, regardless of its context or any other intrinsic property. 
To address the limitations of word2vec and GloVe with out-of-vocabulary words, where morphologically-rich languages such as Finnish or Turkish are specially affected, new models appeared which take subword information into account.
The type of subword information used varies in each particular approach: some of them require a preprocessing step to extract morphemes~\cite{Luong2013}, while others employ a less strict approach by directly using the characters~\cite{Ling2015,Kim2016} or character n-grams~\cite{bojanowski2017enriching,Wieting2016} that form the words.

When targeting noisy texts from social media, such as tweets from Twitter, previous work relies solely on the high coverage that can be obtained from training in an equally noisy domain~\cite{Sumbler2018}.
\mbox{An exception} to this rule is the work from~\citet{Malykh2018}, where they try to obtain robust embeddings to misspelled words (one or two edit operations away from the correct form) by using \mbox{a new} neural-based model.
In this case, the flexibility is obtained by an encoding of the prefix, \mbox{suffix and} set of characters that form each word. 
By using this set of characters in the encoding, \mbox{where the} specific order between them is disregarded, this approach achieves some form of robustness to low-level noise, while the prefix and suffix part encodes most of the semantic information.
The main difference of our approach is that we are not proposing a whole new model but a generic technique to adapt existing ones. 
\mbox{This could} be applied to many others, including that from~\citet{Malykh2018} itself.
\mbox{Furthermore, we evaluate} our embeddings in the context of non-standard texts, a noisier medium than the slightly misspelled standard texts regarded in~\cite{Malykh2018}. 
Unfortunately, we could not include this approach in our test bench as, probably due to differences in the development environment setup, \mbox{we were} not able to train new models nor extract embeddings through pretrained models using the latest version of the code at~\url{https://gitlab.com/madrugado/robust-w2v/tree/py3_launch}.

Lastly, if we consider standard and non-standard texts as pertaining to different languages, \mbox{our approach} would be similar to~\cite{Luong15}, where they also adapt the skipgram model to obtain bilingual embeddings.
In this work, they start with comparable bilingual corpora and automatically calculate alignments between words across languages.
At training time, they use the words from alignment pairs interchangeably in the texts from each language, requiring each word to predict not only the context in its own language but also the context in the other language.
In our case, we only consider one training corpus and create a set of bridge-words that act as alignments between standard words and their noisy counterparts.
On the other hand, the weight given to these new words in the objective function is $\lambda < 1$ as they represent noisy examples, whereas, in~\cite{Luong15}, the words from the other language are given more weight ($\lambda > 1$).

\section{Conclusions}

In this work, we have proposed a modification of the skipgram model from fastText intended to improve the performance of word embedding models on noisy texts as they are found on social media, while retaining the performance on standard texts.
To do this, we introduce a new set of words in the training process, called bridge-words, whose objective is to connect standard words with their \mbox{noisy counterparts.}

We have evaluated the performance of the proposed approach together with word2vec and fastText baselines on a wide array of intrinsic and extrinsic tasks.
The results show that, while the performance of our best model on standard texts is mostly preserved when compared to the baselines, it generally outperforms them on noisier texts with wider margins as the level of noise increases.

As future lines of research, we will perform the same study on other languages and adapt the proposed modification of the skipgram model to work with the newest BERT~\cite{devlin2018bert} and GPT-2~\cite{radford2019language} models, given that GPT-3~\cite{brown2020language} is prohibitively expensive to train.
In light of its competitive performance, adapting the skipgram model from word2vec might prove useful.
Other types of bridge-words such as phonetic codes obtained from a phonetic algorithm like the Metaphone~\cite{philips1990hanging} could also prove to be beneficial, in addition to a weighted term inversely proportional to word length, and \textit{u} or inverted-\textit{u} distributions for noise introduction.
Additionally, our approach is orthogonal to other techniques that enhance the performance of word embeddings, such as the ones described in~\cite{mikolov2017advances}, and so they too can be applied to the models obtained in this work.

%%%%%%%%%%%%%%%%%%%%%%%%%%%%%%%%%%%%%%%%%%
%\section{Patents}
%This section is not mandatory, but may be added if there are patents resulting from the work reported in this manuscript.

%%%%%%%%%%%%%%%%%%%%%%%%%%%%%%%%%%%%%%%%%%
\vspace{6pt} 

%%%%%%%%%%%%%%%%%%%%%%%%%%%%%%%%%%%%%%%%%%
%% optional
%\supplementary{The following are available online at \linksupplementary{s1}, Figure S1: title, Table S1: title, Video S1: title.}


%%%%%%%%%%%%%%%%%%%%%%%%%%%%%%%%%%%%%%%%%%
\authorcontributions{Conceptualization, Y.D.; methodology, Y.D.; software, Y.D.; validation, Y.D. and J.V., \mbox{and C.G.-R.;} formal analysis, Y.D. and C.G.-R.; investigation, Y.D.; resources, Y.D.; data curation, Y.D.; writing---original draft preparation, Y.D.; writing---review and editing, Y.D., J.V., and C.G.-R.; visualization, Y.D. and C.G.-R.; supervision, J.V. and C.G.-R.; project administration, J.V. and C.G.-R.; funding acquisition, J.V. and C.G.-R. All authors have read and agreed to the published version of the manuscript.}

%%%%%%%%%%%%%%%%%%%%%%%%%%%%%%%%%%%%%%%%%%
\funding{Yerai Doval has been supported by the Spanish Ministry of Economy, Industry and Competitiveness (MINECO) through the ANSWER-ASAP project (TIN2017-85160-C2-2-R); by the Spanish State Secretariat for Research, Development and Innovation (which belongs to MINECO) and the European Social Fund (ESF) through a FPI fellowship (BES-2015-073768) associated with TELEPARES project (FFI2014-51978-C2-1-R); and by the Xunta de Galicia through TELGALICIA research network (ED431D 2017/12). 
The work of Jes\'us Vilares and Carlos G\'omez-Rodr\'iguez has also been funded by MINECO through the ANSWER-ASAP project (TIN2017-85160-C2-1-R in this case); and by Xunta de Galicia through a Group with Potential for Growth grant (ED431B 2017/01), a Competitive Reference Group grant (ED431C 2020/11), and a Remarkable Research Centre grant for the CITIC research centre (ED431G/01), the latter co-funded by EU with ERDF funding. \mbox{Finally, Carlos G\'omez-Rodr\'iguez}  has also received funding from the European Research Council (ERC), under the European Union's Horizon 2020 research and innovation programme (FASTPARSE, Grant No. 714150).}

%%%%%%%%%%%%%%%%%%%%%%%%%%%%%%%%%%%%%%%%%%
%\acknowledgments{In this section you can acknowledge any support given which is not covered by the author contribution or funding sections. This may include administrative and technical support, or donations in kind (e.g., materials used for experiments).}

%%%%%%%%%%%%%%%%%%%%%%%%%%%%%%%%%%%%%%%%%%
\conflictsofinterest{The authors declare no conflict of interest.} 

%%%%%%%%%%%%%%%%%%%%%%%%%%%%%%%%%%%%%%%%%%
%% optional
%\abbreviations{The following abbreviations are used in this manuscript:\\

%\noindent 
%\begin{tabular}{@{}ll}
%SA & Sentiment Analysis\\
%LD & linear dichroism
%\end{tabular}}

%%%%%%%%%%%%%%%%%%%%%%%%%%%%%%%%%%%%%%%%%%
%% optional
\appendixtitles{yes} %Leave argument "no" if all appendix headings stay EMPTY (then no dot is printed after "Appendix A"). If the appendix sections contain a heading then change the argument to "yes".
\appendix
\section{The Importance of Word Segmentation}

In principle, when using word embeddings, we assume that the input text is correctly segmented into words.
However, suppose that this is not the case, and that it goes beyond frequent de-normalization instances where words are joined or merged together; e.g., `noway'-`no way', `yesplease'-`yes please'; or even instances where the individual words cannot be immediately recovered such as with `tryna'-`trying to', or `whatchu'-`what are/do you'. 
Instead, in the present case, we will consider sentences like `theproblem was veryclear' or `the prob lem was very clear'. 
A possible solution would be to perform a word segmentation preprocessing step~\cite{doval2019comparing} before obtaining the corresponding word embeddings, which would imply introducing again the notion of sequential tasks together with the risk of error propagation.%, as explained in Section~\ref{sec:inter.limitations}.

However, let us now consider that the two operations involved in bad word segmentation \mbox{(i.e., word joining} and splitting) might not have the same impact on the process of obtaining relevant word embeddings. 
If we take into account that models such as fastText, and by extension the modification presented in this chapter, use subword information to construct word embeddings, we might argue that \textit{joining words together} may be moderately supported by these models, as they would still consider the words inside the merging as character n-grams modelled during training. 
\mbox{On the contrary,} \textit{\mbox{splitting words}} would be more problematic, as it removes parts of a word which could be crucial to obtain the adequate vector representation. 
%Hence, we also expect that models such as word2vec should be in clear disadvantage in this case.

To check this hypothesis, we have devised new experiments using new de-normalized versions of the STS* datasets from the SentEval benchmark, which we have divided into two sets: \textit{join} and \textit{split}.
In the former, we randomly removed word delimiters from input sentences with a fixed probability $p_j$, while in the latter we added delimiters between word characters with a lower probability $p_s$, $p_s < p_j$, in order to account for the higher amount of non-delimiter characters.

The results obtained, which are shown in Table~\ref{table:joinsplit}, seem to support our hypothesis.
\mbox{Therefore, using a} word segmenter with a slight tendency to join words (e.g., through a threshold parameter as shown by~\citet{doval2016segmentacion}) or even the raw input directly (taking into account the low frequency of splits, while joins are frequent in special elements such as hashtags or URLs), \mbox{can be} considered good practical solutions so long as we use embedding models that exploit subword information.
Nonetheless, the latter option is especially relevant for us, since it shows that we may finally dispense with any form of input preprocessing for languages that delimit words; English in our current case.
However, even in the case of Chinese, where words are not explicitly delimited and word segmentation is a well-studied and complex subject, it has been recently shown that this preprocessing step might not be necessary.
\citet{meng2019word} propose directly operating over Chinese characters rather than \textit{strict} words.
We highlight this strictness property as characters are frequently used as words themselves, but not always. 
That solution obtains better results than other systems that require a previous word segmentation step, even when all of them are implemented as state-of-the-art neural networks.
\mbox{In our case,} this shows that we could relax the definition of a \textit{word}, and obtain the embeddings at the character- or sequence-of-characters level.

\begin{table}[H]
\centering
\caption{Spearman correlation averages on the new de-normalized STS* datasets, with $p_j = 0.5$ and $p_s=0.1$.} 
\begin{tabular}{ccc}
\toprule
         & \multicolumn{1}{c}{\textbf{Join\boldmath{$_{\rho}$}}} & \multicolumn{1}{c}{\textbf{Split\boldmath{$_{\rho}$}}} \\ \midrule
word2vec      & 11.0                    & 18.3                     \\
fastText & \textbf{39.2}                    & \textbf{18.7}                     \\
ours     & \textbf{39.2}                    & 17.2      \\
\bottomrule
\end{tabular}
\label{table:joinsplit}
\end{table}

%\unskip
%\subsection{}
%The appendix is an optional section that can contain details and data supplemental to the main text. For example, explanations of experimental details that would disrupt the flow of the main text, but nonetheless remain crucial to understanding and reproducing the research shown; figures of replicates for experiments of which representative data are shown in the main text can be added here if brief, or as Supplementary data. Mathematical proofs of results not central to the paper can be added as an appendix.

%\section{}
%All appendix sections must be cited in the main text. In the appendixes, Figures, Tables, etc. should be labeled starting with `A', e.g., Figure A1, Figure A2, etc. 


%=====================================
% References, variant A: internal bibliography
%=====================================
\reftitle{References}

%\begin{thebibliography}{999}
% Reference 1
%\bibitem[Author1(year)]{ref-journal}
%Author1, T. The title of the cited article. {\em Journal Abbreviation} {\bf 2008}, {\em 10}, 142--149.
% Reference 2
%\bibitem[Author2(year)]{ref-book}
%Author2, L. The title of the cited contribution. In {\em The Book Title}; Editor1, F., Editor2, A., Eds.; Publishing House: City, Country, 2007; pp. 32--58.
%\end{thebibliography}

% The following MDPI journals use author-date citation: Arts, Econometrics, Economies, Genealogy, Humanities, IJFS, JRFM, Laws, Religions, Risks, Social Sciences. For those journals, please follow the formatting guidelines on http://www.mdpi.com/authors/references
% To cite two works by the same author: \citeauthor{ref-journal-1a} (\citeyear{ref-journal-1a}, \citeyear{ref-journal-1b}). This produces: Whittaker (1967, 1975)
% To cite two works by the same author with specific pages: \citeauthor{ref-journal-3a} (\citeyear{ref-journal-3a}, p. 328; \citeyear{ref-journal-3b}, p.475). This produces: Wong (1999, p. 328; 2000, p. 475)

%=====================================
% References, variant B: external bibliography
%=====================================
%\externalbibliography{yes}
%\bibliography{biblio}

\begin{thebibliography}{999}
%\providecommand{\natexlab}[1]{#1}

\bibitem[Bansal et~al.(2014)Bansal, Gimpel, and Livescu]{Bansal2014}
Bansal, M.; Gimpel, K.; Livescu, K.
\newblock Tailoring Continuous Word Representations for Dependency Parsing.
\newblock In Proceedings of the 52nd Annual Meeting of the Association for
  Computational Linguistics (\mbox{Volume 2:} Short Papers), Baltimore, MD, USA, 22--27  June 2014;  Association for
  Computational Linguistics: \mbox{Baltimore, MD, USA,}  2014; pp. 809--815,
\newblock
  doi:{\changeurlcolor{black}\href{https://doi.org/10.3115/v1/P14-2131}{\detokenize{10.3115/v1/P14-2131}}}.

\bibitem[Vuli{\'{c}} and Moens(2015)]{vulic2015monolingual}
Vuli{\'{c}}, I.; Moens, M.F.
\newblock {Monolingual and Cross-Lingual Information Retrieval Models Based on
  (Bilingual) Word Embeddings}.
\newblock In Proceedings of the 38th International ACM SIGIR Conference on
  Research and Development in Information Retrieval---SIGIR '15, Santiago, Chile, 11--15 August  2015;
  pp. 363--372,
\newblock
  doi:{\changeurlcolor{black}\href{https://doi.org/10.1145/2766462.2767752}{\detokenize{10.1145/2766462.2767752}}}.

\bibitem[Kutuzov et~al.(2016)Kutuzov, Velldal, and
  {\O}vrelid]{Kutuzov2016}
Kutuzov, A.; Velldal, E.; {\O}vrelid, L.
\newblock Redefining part-of-speech classes with distributional semantic
  models.
\newblock  In  Proceedings of the 20th {SIGNLL} Conference on Computational Natural
  Language Learning, Berlin, Germany, 11--12 August 2016; Association for Computational Linguistics: \mbox{Berlin,
  Germany,}  2016; \mbox{pp. 115--125,}
\newblock
  doi:{\changeurlcolor{black}\href{https://doi.org/10.18653/v1/K16-1012}{\detokenize{10.18653/v1/K16-1012}}}.

\bibitem[Xiong et~al.(2018)Xiong, Lv, Zhao, and Ji]{Xiong2018}
Xiong, S.; Lv, H.; Zhao, W.; Ji, D.
\newblock {Towards Twitter sentiment classification by multi-level
  sentiment-enriched word embeddings}.
\newblock {\em Neurocomputing} {\bf 2018}, {\em 275},~2459--2466,
\newblock
  doi:{\changeurlcolor{black}\href{https://doi.org/10.1016/j.neucom.2017.11.023}{\detokenize{10.1016/j.neucom.2017.11.023}}}.

\bibitem[Lampos et~al.(2017)Lampos, Zou, and Cox]{Lampos2017}
Lampos, V.; Zou, B.; Cox, I.J.
\newblock Enhancing Feature Selection Using Word Embeddings: The Case of Flu
  Surveillance.
\newblock In Proceedings of the 26th International Conference on World Wide Web,
  Perth, Australia, \mbox{3--7 May 2017;} {ACM}: Perth, Australia,  2017; pp. 695--704,
\newblock
  doi:{\changeurlcolor{black}\href{https://doi.org/10.1145/3038912.3052622}{\detokenize{10.1145/3038912.3052622}}}.

\bibitem[Yang et~al.(2018)Yang, Macdonald, and Ounis]{Yang2018}
Yang, X.; Macdonald, C.; Ounis, I.
\newblock {Using word embeddings in Twitter election classification}.
\newblock {\em Inf. Retr. J.} {\bf 2018}, {\em 21},~183--207,
\newblock
  doi:{\changeurlcolor{black}\href{https://doi.org/10.1007/s10791-017-9319-5}{\detokenize{10.1007/s10791-017-9319-5}}}.

\bibitem[Liang et~al.(2018)Liang, Zhang, Ren, and Kanoulas]{Liang2018}
Liang, S.; Zhang, X.; Ren, Z.; Kanoulas, E.
\newblock {Dynamic Embeddings for User Profiling in Twitter}.
\newblock In Proceedings of the 24th {ACM} {SIGKDD} International Conference on
  Knowledge Discovery {\&} Data Mining,  \mbox{{KDD}, London, UK,} 19--23 August  2018; {ACM}: London, UK,  2018;
  pp. 1764--1773,
\newblock
  doi:{\changeurlcolor{black}\href{https://doi.org/10.1145/3219819.3220043}{\detokenize{10.1145/3219819.3220043}}}.

\bibitem[Mikolov et~al.(2013)Mikolov, Corrado, Chen, and
  Dean]{Mikolov2013}
Mikolov, T.; Corrado, G.; Chen, K.; Dean, J.
\newblock Efficient estimation of word representations in vector space.
\newblock In Proceedings of the 1st International Conference on Learning
  Representations, Scottsdale, AZ, USA, \mbox{2--4 May 2013;} pp. 1--12,
\newblock
  doi:{\changeurlcolor{black}\href{https://doi.org/10.1037/pspa0000033}{\detokenize{10.1037/pspa0000033}}}.

\bibitem[Pennington et~al.(2014)Pennington, Socher, and
  Manning]{Pennington2014}
Pennington, J.; Socher, R.; Manning, C.
\newblock {G}lo{V}e: Global Vectors for Word Representation.
\newblock  In Proceedings of the 2014 Conference on Empirical Methods in Natural
  Language Processing ({EMNLP}),  Doha, Qatar, 25--29 October  2014; Association for Computational Linguistics:
  Doha, Qatar,  2014; pp. 1532--1543,
\newblock
  doi:{\changeurlcolor{black}\href{https://doi.org/10.3115/v1/D14-1162}{\detokenize{10.3115/v1/D14-1162}}}.

\bibitem[Bojanowski et~al.(2017)Bojanowski, Grave, Joulin, and
  Mikolov]{bojanowski2017enriching}
Bojanowski, P.; Grave, E.; Joulin, A.; Mikolov, T.
\newblock {Enriching Word Vectors with Subword Information}.
\newblock {\em \mbox{Trans. Assoc.} Comput. Linguist.}
  {\bf 2017}, {\em 5},~135--146,
\newblock
  doi:{\changeurlcolor{black}\href{https://doi.org/10.1007/BF01959819}{\detokenize{10.1007/BF01959819}}}.

\bibitem[Sumbler et~al.(2018)Sumbler, Viereckel, Afsarmanesh, and
  Karlgren]{Sumbler2018}
Sumbler, P.; Viereckel, N.; Afsarmanesh, N.; Karlgren, J.
\newblock Handling Noise in Distributional Semantic Models for Large Scale Text
  Analytics and Media Monitoring.
\newblock  In Proceedings of the Abstract in the Fourth Workshop on Noisy User---Generated Text
  (W-NUT 2018), Brussels, Belgium, 1 November 2018.

\bibitem[Eisenstein(2013)]{Eisenstein2013}
Eisenstein, J.
\newblock What to do about bad language on the internet.
\newblock In  Proceedings of the 2013 Conference of the North {A}merican Chapter
  of the Association for Computational Linguistics: Human Language
  Technologies, Atlanta, GA, USA,  9--14 June 2013; Association for Computational Linguistics: Atlanta, GA, USA,
  2013; \mbox{pp. 359--369.}

\bibitem[Chrupa{\l}a(2014)]{Chrupaa2014}
Chrupa{\l}a, G.
\newblock Normalizing tweets with edit scripts and recurrent neural embeddings.
\newblock In  Proceedings of the 52nd Annual Meeting of the Association for
  Computational Linguistics (Volume 2: Short Papers), \mbox{Baltimore, MD, USA,} 22--27 June 2014; Association for
  Computational Linguistics: Baltimore, MD, USA,  2014; pp. 680--686,
\newblock
  doi:{\changeurlcolor{black}\href{https://doi.org/10.3115/v1/P14-2111}{\detokenize{10.3115/v1/P14-2111}}}.

\bibitem[Bordes et~al.(2017)Bordes, Boureau, and Weston]{Bordes2017}
Bordes, A.; Boureau, Y.; Weston, J.
\newblock {Learning End-to-End Goal-Oriented Dialog}.
\newblock In  Proceedings of the  5th International Conference on Learning Representations, 
   Toulon, France,  24--26 April  2017.

\bibitem[Klein et~al.(2017)Klein, Kim, Deng, Senellart, and
  Rush]{Klein2017}
Klein, G.; Kim, Y.; Deng, Y.; Senellart, J.; Rush, A.
\newblock {O}pen{NMT}: Open-Source Toolkit for Neural Machine Translation.
\newblock In Proceedings of {ACL} 2017, System Demonstrations, Vancouver, BC, Canada, \mbox{30 July--4 August 2017;} Association for  Computational Linguistics: Vancouver, BC, Canada,  2017; pp. 67--72.

\bibitem[Schmitt et~al.(2018)Schmitt, Steinheber, Schreiber, and
  Roth]{Schmitt2018}
Schmitt, M.; Steinheber, S.; Schreiber, K.; Roth, B.
\newblock Joint Aspect and Polarity Classification for Aspect-based Sentiment
  Analysis with End-to-End Neural Networks.
\newblock In Proceedings of the 2018 Conference on Empirical Methods in Natural
  Language Processing, Brussels, Belgium, 31 October--4 November 2018; pp. 1109--1114.

\bibitem[van~der Goot et~al.(2017)van~der Goot, Plank, and
  Nissim]{Van2017a}
van~der Goot, R.; Plank, B.; Nissim, M.
\newblock To normalize, or not to normalize: The impact of normalization on
  Part-of-Speech tagging.
\newblock In Proceedings of the 3rd Workshop on Noisy User-generated Text,  Copenhagen, Denmark, 7 September 2017;  pp. 31--39.

\bibitem[Costa~Bertaglia and Volpe~Nunes(2016)]{Bertaglia2017}
Costa~Bertaglia, T.F.; Volpe~Nunes, M.D.G.
\newblock Exploring Word Embeddings for Unsupervised Textual User-Generated
  Content Normalization.
\newblock In Proceedings of the 2nd Workshop on Noisy User-generated Text
  ({WNUT}), Osaka, Japan, 19 November 2020; The COLING 2016 Organizing Committee: Osaka, Japan,  2016; pp.
  112--120.

\bibitem[van~der Goot and van Noord(2017)]{Van2017b}
Van~der Goot, R.; van Noord, G.
\newblock MoNoise: Modeling Noise Using a Modular Normalization System. 
\newblock {\em \mbox{Comput. Linguist.} Neth. J.} {\bf
  2017}, {\em 7},~129--144.

\bibitem[Ansari et~al.(2017)Ansari, Zafar, and Karim]{Ansari2017}
Ansari, S.A.; Zafar, U.; Karim, A.
\newblock Improving Text Normalization by Optimizing Nearest Neighbor Matching.
\newblock {\em arXiv} {\bf 2017}, arXiv:1712.09518.

\bibitem[Sridhar(2015)]{sridhar2015unsupervised}
Sridhar, V.K.R.
\newblock Unsupervised text normalization using distributed representations of
  words and phrases.
\newblock In Proceedings of the 1st Workshop on Vector Space Modeling for Natural
  Language Processing, \mbox{Denver, CO, USA,} 31 May--5 June  2015; pp. 8--16.

\bibitem[Devlin et~al.(2019)Devlin, Chang, Lee, and
  Toutanova]{devlin2018bert}
Devlin, J.; Chang, M.; Lee, K.; Toutanova, K.
\newblock {BERT:} Pre-training of Deep Bidirectional Transformers for Language
  Understanding.
\newblock In Proceedings of the 2019 Conference of the North American Chapter of
  the Association for Computational Linguistics: Human Language Technologies,
  \mbox{{NAACL-HLT} 2019,} \mbox{Minneapolis, MN, USA,} 2--7 June  2019;  Burstein, J., Doran, C., Solorio, T., Eds.; Association for
  Computational Linguistics:   Minneapolis, MN, USA, 2019; Volume 1 (Long and
  Short Papers), pp. 4171--4186,
\newblock
  doi:{\changeurlcolor{black}\href{https://doi.org/10.18653/v1/n19-1423}{\detokenize{10.18653/v1/n19-1423}}}.

\bibitem[Radford et~al.(2019)Radford, Wu, Child, Luan, Amodei, and
  Sutskever]{radford2019language}
Radford, A.; Wu, J.; Child, R.; Luan, D.; Amodei, D.; Sutskever, I.
\newblock Language models are unsupervised multitask learners.
\newblock {\em OpenAI Blog} {\bf 2019}, {\em 1},~9.

\bibitem[Brown et~al.(2020)Brown, Mann, Ryder, Subbiah, Kaplan, Dhariwal,
  Neelakantan, Shyam, Sastry, Askell, Agarwal, Herbert{-}Voss, Krueger,
  Henighan, Child, Ramesh, Ziegler, Wu, Winter, Hesse, Chen, Sigler, Litwin,
  Gray, Chess, Clark, Berner, McCandlish, Radford, Sutskever, and
  Amodei]{brown2020language}
Brown, T.B.; Mann, B.; Ryder, N.; Subbiah, M.; Kaplan, J.; Dhariwal, P.;
  Neelakantan, A.; Shyam, P.; Sastry, G.; Askell, A.; et al.
\newblock Language Models are Few-Shot Learners.
\newblock {\em arXiv } {\bf 2020}, arXiv:2005.14165.

\bibitem[Camacho-Collados and Navigli(2016)]{camacho2016find}
Camacho-Collados, J.; Navigli, R.
\newblock Find the word that does not belong: A Framework for an Intrinsic
  Evaluation of Word Vector Representations.
\newblock  In Proceedings of the 1st Workshop on Evaluating Vector-Space
  Representations for {NLP},  Berlin,
  Germany, 12 August 2016; Association for Computational Linguistics: Berlin,
  Germany,  2016; pp. 43--50,
\newblock
  doi:{\changeurlcolor{black}\href{https://doi.org/10.18653/v1/W16-2508}{\detokenize{10.18653/v1/W16-2508}}}.

\bibitem[Conneau and Kiela(2018)]{conneau2018senteval}
Conneau, A.; Kiela, D.
\newblock {S}ent{E}val: An Evaluation Toolkit for Universal Sentence
  Representations.
\newblock \mbox{In Proceedings} of the Eleventh International Conference on Language
  Resources and Evaluation ({LREC} 2018), Miyazaki, Japan, 7--12 May 2018; European Language Resources
  Association (ELRA): \mbox{Miyazaki, Japan,  2018.}

\bibitem[Han et~al.(2013)Han, L.~Kashyap, Finin, Mayfield, and
  Weese]{UMBC}
Han, L.; Kashyap, A.L.; Finin, T.; Mayfield, J.; Weese, J.
\newblock {UMBC}{\_}{EBIQUITY}-{CORE}: Semantic Textual Similarity Systems.
\newblock  In \emph{Second Joint Conference on Lexical and Computational Semantics
  (*{SEM}), Volume 1: Proceedings of the Main Conference and the Shared Task:
  Semantic Textual Similarity}; Association for Computational Linguistics:
  Atlanta, GA, USA,  2013; pp. 44--52.

\bibitem[Finkelstein et~al.(2002)Finkelstein, Evgeniy, Yossi, Ehud, Zach,
  Gadi, and Eytan]{Levetal:2002}
Finkelstein, L.; Evgeniy, G.; Yossi, M.; Ehud, R.; Zach, S.; Gadi, W.; Eytan,
  R.
\newblock Placing Search in Context: \mbox{The Concept} Revisited.
\newblock {\em ACM Trans. Inf. Syst.} {\bf 2002}, {\em
  20},~116--131.

\bibitem[Huang et~al.(2012)Huang, Socher, Manning, and
  Ng]{huang2012improving}
Huang, E.; Socher, R.; Manning, C.; Ng, A.
\newblock Improving Word Representations via Global Context and Multiple Word
  Prototypes.
\newblock In Proceedings of the 50th Annual Meeting of the Association for
  Computational Linguistics (Volume 1: Long Papers), Jeju Island, Korea, 8--14 July 2012; Association for
  Computational Linguistics: Jeju Island, Korea,  2012; pp. 873--882.

\bibitem[Hill et~al.(2015)Hill, Reichart, and Korhonen]{hill2015simlex}
Hill, F.; Reichart, R.; Korhonen, A.
\newblock {S}im{L}ex-999: Evaluating Semantic Models With (Genuine) Similarity
  Estimation.
\newblock {\em Comput. Linguist.} {\bf 2015}, {\em 41},~665--695,
\newblock
  doi:{\changeurlcolor{black}\href{https://doi.org/10.1162/COLI_a_00237}{\detokenize{10.1162/COLI_a_00237}}}.

\bibitem[Camacho-Collados et~al.(2017)Camacho-Collados, Pilehvar, Collier,
  and Navigli]{semeval2017similarity}
Camacho-Collados, J.; Pilehvar, M.T.; Collier, N.; Navigli, R.
\newblock {S}em{E}val-2017 Task 2: Multilingual and Cross-lingual Semantic Word
  Similarity.
\newblock In Proceedings of the 11th International Workshop on Semantic
  Evaluation ({S}em{E}val-2017), Vancouver, BC, Canada, 3--4 August 2017; Association for Computational Linguistics:
  Vancouver, BC, Canada,  2017; pp. 15--26,
\newblock
  doi:{\changeurlcolor{black}\href{https://doi.org/10.18653/v1/S17-2002}{\detokenize{10.18653/v1/S17-2002}}}.

\bibitem[Blair et~al.(2017)Blair, Merhav, and Barry]{blair2016automated}
Blair, P.; Merhav, Y.; Barry, J.
\newblock Automated Generation of Multilingual Clusters for the Evaluation of
  Distributed Representations.
\newblock In Proceedings of the 5th International Conference on Learning Representations, \mbox{{ICLR}
  2017,} Toulon, France,  24--26 April 2017.

\bibitem[Faruqui et~al.(2016)Faruqui, Tsvetkov, Rastogi, and
  Dyer]{faruqui2016problems}
Faruqui, M.; Tsvetkov, Y.; Rastogi, P.; Dyer, C.
\newblock Problems With Evaluation of Word Embeddings Using Word Similarity
  Tasks.
\newblock  In Proceedings of the 1st Workshop on Evaluating Vector-Space
  Representations for {NLP}, Berlin,   Germany,  12 August 2016; Association for Computational Linguistics: Berlin,
  Germany,  2016; \mbox{pp. 30--35,}
\newblock
  doi:{\changeurlcolor{black}\href{https://doi.org/10.18653/v1/W16-2506}{\detokenize{10.18653/v1/W16-2506}}}.

\bibitem[Chiu et~al.(2016)Chiu, Korhonen, and Pyysalo]{chiu2016intrinsic}
Chiu, B.; Korhonen, A.; Pyysalo, S.
\newblock Intrinsic Evaluation of Word Vectors Fails to Predict Extrinsic
  Performance.
\newblock  In Proceedings of the 1st Workshop on Evaluating Vector-Space
  Representations for {NLP}, Berlin,   Germany,  12 August 2016; Association for Computational Linguistics: Berlin,
  Germany,  2016; pp. 1--6,
\newblock
  doi:{\changeurlcolor{black}\href{https://doi.org/10.18653/v1/W16-2501}{\detokenize{10.18653/v1/W16-2501}}}.

\bibitem[Conneau et~al.(2017)Conneau, Kiela, Schwenk, Barrault, and
  Bordes]{conneau2017supervised}
Conneau, A.; Kiela, D.; Schwenk, H.; Barrault, L.; Bordes, A.
\newblock Supervised Learning of Universal Sentence Representations from
  Natural Language Inference Data.
\newblock In  Proceedings of the 2017 Conference on Empirical Methods in Natural
  Language Processing, Copenhagen, Denmark, 7--11 September 2017; Association for Computational Linguistics: Copenhagen,
  Denmark,  2017; pp. 670--680.

\bibitem[Subramanian et~al.(2018)Subramanian, Trischler, Bengio, and
  Pal]{subramanian2018learning}
Subramanian, S.; Trischler, A.; Bengio, Y.; Pal, C.J.
\newblock Learning General Purpose Distributed Sentence Representations via
  Large Scale Multi-task Learning.
\newblock In Proceedings of the International Conference on Learning Representations, Vancouver, BC, Canada,  30 April--3 May 2018; pp.
  1--16.

\bibitem[Nakov et~al.(2013)Nakov, Rosenthal, Kozareva, Stoyanov, Ritter,
  and Wilson]{nakov2013semeval}
Nakov, P.; Rosenthal, S.; Kozareva, Z.; Stoyanov, V.; Ritter, A.; Wilson, T.
\newblock Sem{E}val-2013 Task 2: Sentiment Analysis in {T}witter.
\newblock  In Proceedings of the 7th International Workshop on Semantic
  Evaluation, SemEval@NAACL-HLT 2013, Atlanta, GA, USA, 14--15 June 2013; Association for Computational
  Linguistics: Atlanta, GA, USA,  2013; pp. 312--320.

\bibitem[Rosenthal et~al.(2014)Rosenthal, Ritter, Nakov, and
  Stoyanov]{rosenthal2014semeval}
Rosenthal, S.; Ritter, A.; Nakov, P.; Stoyanov, V.
\newblock SemEval-2014 Task 9: Sentiment Analysis in {T}witter.
\newblock   \mbox{In Proceedings} of the 8th International Workshop on Semantic
  Evaluation, SemEval@COLING 2014, \mbox{Dublin, Ireland,}  23--24 August 2014; pp. 73--80.

\bibitem[Nakov et~al.(2016)Nakov, Ritter, Rosenthal, Sebastiani, and
  Stoyanov]{nakov2016semeval}
Nakov, P.; Ritter, A.; Rosenthal, S.; Sebastiani, F.; Stoyanov, V.
\newblock {S}em{E}val-2016 Task 4: Sentiment Analysis in {T}witter.
\newblock  In \emph{ Proceedings of the 10th International Workshop on Semantic
  Evaluation ({S}em{E}val-2016)}; Association for Computational Linguistics: San
  Diego, CA, USA,  2016; pp. 1--18,
\newblock
  doi:{\changeurlcolor{black}\href{https://doi.org/10.18653/v1/S16-1001}{\detokenize{10.18653/v1/S16-1001}}}.

\bibitem[Baldwin et~al.(2015)Baldwin, de~Marneffe, Han, Kim, Ritter, and
  Xu]{baldwin2015shared}
Baldwin, T.; de~Marneffe, M.C.; Han, B.; Kim, Y.B.; Ritter, A.; Xu, W.
\newblock Shared tasks of the 2015 workshop on noisy user-generated text:
  Twitter lexical normalization and named entity recognition.
\newblock In Proceedings of the Workshop on Noisy User----Generated Text, Beijing, China, 31 July 2015; pp.
  126--135.

\bibitem[Rumelhart et~al.(1986)Rumelhart, Hinton, and
  Williams]{Rumelhart1986}
Rumelhart, D.E.; Hinton, G.E.; Williams, R.J.
\newblock Learning representations by back-propagating errors.
\newblock \mbox{{\em Nature} {\bf 1986},} {\em 323},~533.

\bibitem[Luong et~al.(2013)Luong, Socher, and Manning]{Luong2013}
Luong, T.; Socher, R.; Manning, C.
\newblock Better Word Representations with Recursive Neural Networks for
  Morphology.
\newblock In Proceedings of the 17th Conference on Computational Natural Language
  Learning, \mbox{Sofia, Bulgaria,} 8--9 August 2013; Association for Computational Linguistics: Sofia, Bulgaria,  2013;  \mbox{pp. 104--113.}

\bibitem[Ling et~al.(2015)Ling, Dyer, Black, Trancoso, Fermandez, Amir,
  Marujo, and Lu{\'\i}s]{Ling2015}
Ling, W.; Dyer, C.; Black, A.W.; Trancoso, I.; Fermandez, R.; Amir, S.; Marujo,
  L.; Lu{\'\i}s, T.
\newblock Finding Function in Form: Compositional Character Models for Open
  Vocabulary Word Representation.
\newblock \mbox{In  Proceedings} of the 2015 Conference on Empirical Methods in Natural
  Language Processing, Lisbon,
  Portugal,  5 June 2015;  Association for Computational Linguistics: Lisbon,
  Portugal,  2015; pp. 1520--1530,
\newblock
  doi:{\changeurlcolor{black}\href{https://doi.org/10.18653/v1/D15-1176}{\detokenize{10.18653/v1/D15-1176}}}.

\bibitem[Kim et~al.(2016)Kim, Jernite, Sontag, and Rush]{Kim2016}
Kim, Y.; Jernite, Y.; Sontag, D.A.; Rush, A.M.
\newblock Character-Aware Neural Language Models.
\newblock In Proceedings of the Thirtieth {AAAI} Conference on Artificial
  Intelligence, Phoenix, AZ, USA, 12--17 February 2016; \mbox{{AAAI} Press:} Phoenix, AZ, {USA},  2016; pp. 2741--2749.

\bibitem[Wieting et~al.(2016)Wieting, Bansal, Gimpel, and
  Livescu]{Wieting2016}
Wieting, J.; Bansal, M.; Gimpel, K.; Livescu, K.
\newblock {C}haragram: Embedding Words and Sentences via Character n-grams.
\newblock In Proceedings of the 2016 Conference on Empirical Methods in Natural
  Language Processing, Austin,
  TX, USA, 1--4 November 2016; Association for Computational Linguistics: Austin,
  TX, USA,  2016; pp. 1504--1515,
\newblock
  doi:{\changeurlcolor{black}\href{https://doi.org/10.18653/v1/D16-1157}{\detokenize{10.18653/v1/D16-1157}}}.

\bibitem[Malykh et~al.(2018)Malykh, Logacheva, and Khakhulin]{Malykh2018}
Malykh, V.; Logacheva, V.; Khakhulin, T.
\newblock Robust Word Vectors: Context-Informed Embeddings for Noisy Texts.
\newblock In  Proceedings of the 2018 EMNLP Workshop W-NUT: The 4th Workshop on
  Noisy User---Generated Text, Brussels, Belgium, 9--15 September  2018; pp. 54--63.

\bibitem[Luong et~al.(2015)Luong, Pham, and Manning]{Luong15}
Luong, T.; Pham, H.; Manning, C.D.
\newblock Bilingual Word Representations with Monolingual Quality in Mind.
\newblock In Proceedings of the 1st Workshop on Vector Space Modeling for Natural
  Language Processing, \mbox{Denver,
  CO, USA,} 5 June 2015; Association for Computational Linguistics: Denver,
  CO,  USA, 2015; pp. 151--159,
\newblock
  doi:{\changeurlcolor{black}\href{https://doi.org/10.3115/v1/W15-1521}{\detokenize{10.3115/v1/W15-1521}}}.

\bibitem[Philips(1990)]{philips1990hanging}
Philips, L.
\newblock Hanging on the metaphone.
\newblock {\em Comput. Sci.} {\bf 1990}, {\em 7}, 39--43.

\bibitem[Mikolov et~al.(2018)Mikolov, Grave, Bojanowski, Puhrsch, and
  Joulin]{mikolov2017advances}
Mikolov, T.; Grave, E.; Bojanowski, P.; Puhrsch, C.; Joulin, A.
\newblock Advances in Pre-Training Distributed Word Representations.
\newblock In Proceedings of the 11th International Conference on Language
  Resources and Evaluation ({LREC} 2018), Miyazaki, Japan,  7--12 May  2018;  European Language Resources
  Association (ELRA): \mbox{Miyazaki, Japan,}  2018.

\bibitem[Doval and G{\'o}mez-Rodr{\'\i}guez(2019)]{doval2019comparing}
Doval, Y.; G{\'o}mez-Rodr{\'\i}guez, C.
\newblock Comparing neural-and N-gram-based language models for word
  segmentation.
\newblock {\em J. Assoc. Inf. Sci. Technol.} {\bf 2019}, {\em 70},~187--197.

\bibitem[Doval et~al.(2016)Doval, G\'omez-Rodr\'{\i}guez, and
  Vilares]{doval2016segmentacion}
Doval, Y.; G\'omez-Rodr\'{\i}guez, C.; Vilares, J.
\newblock Spanish word segmentation through neural language models.
\newblock {\em Proces. Del Leng. Nat.} {\bf 2016}, {\em
  57},~75--82.

\bibitem[Meng et~al.(2019)Meng, Li, Sun, Han, Yuan, and Li]{meng2019word}
Meng, Y.; Li, X.; Sun, X.; Han, Q.; Yuan, A.; Li, J.
\newblock {Is Word Segmentation Necessary for Deep Learning of Chinese
  Representations?}
\newblock In Proceedings of the 57th Conference of the Association for
  Computational Linguistics, Florence, Italy, 28 July--2 August  2019;  pp. 3242--3252.

\end{thebibliography}

%%%%%%%%%%%%%%%%%%%%%%%%%%%%%%%%%%%%%%%%%%
%% optional
%\sampleavailability{Samples of the compounds ...... are available from the authors.}

%% for journal Sci
%\reviewreports{\\
%Reviewer 1 comments and authors’ response\\
%Reviewer 2 comments and authors’ response\\
%Reviewer 3 comments and authors’ response
%}

%%%%%%%%%%%%%%%%%%%%%%%%%%%%%%%%%%%%%%%%%%
\end{document}